%% file: main.tex
\documentclass[lettersize,journal]{IEEEtran}
\usepackage{amsmath}
\usepackage{amssymb}
\usepackage{amsfonts}
\usepackage{chemarrow}
\usepackage{makecell}
\usepackage{float}
\usepackage{hyperref}
\usepackage{graphicx}
\usepackage{subfigure}
\usepackage{url}
\usepackage{xspace}
\usepackage{color}
\usepackage[linesnumbered,ruled]{algorithm2e}
\newcommand{\BfPara}[1]{{\noindent\bf#1.}\xspace}
\usepackage{multirow, multicol}
\usepackage{float}
\usepackage{supertabular,booktabs}
\usepackage{hyperref}

\usepackage[capitalise]{cleveref}

\usepackage{framed}
\usepackage[T1]{fontenc}
\usepackage[utf8]{inputenc}

\usepackage{environ}
\usepackage{tikz}
\usetikzlibrary{calc,matrix}
{\bfseries}{\itshape}
{\bfseries}{\itshape}
{\bfseries}{\itshape}
{\bfseries}{\itshape}
{\bfseries}{\itshape}
{\itshape}

\usepackage[inline]{enumitem}
\usepackage{caption}

\usepackage{xspace}
\usepackage{setspace}
\newcommand{\note}[1]{}

\renewcommand{\baselinestretch}{1}

\newcommand{\etc}{{etc.}\xspace}

\newcommand{\ie}{{\em i.e.,}\xspace}

\definecolor{linkcolour}{rgb}{0,0.2,0.6}
\definecolor{xgreen}{rgb}{0.2,0.6,0.0}
\definecolor{xred}{rgb}{0.7,0.1,0.0}
\hypersetup{colorlinks,breaklinks,citecolor=green!30!black, 
                                            urlcolor=red!20!black, 
                                            linkcolor=blue!50!black}
\def\equationautorefname~#1\null{(#1)\null}

\usepackage{tikz}
\usetikzlibrary{shapes.geometric, arrows}
\tikzstyle{startstop} = [rectangle, rounded corners, minimum width=3cm, minimum height=1cm,text centered, draw=black, fill=red!30]
\tikzstyle{io} = [trapezium, trapezium left angle=70, trapezium right angle=110, minimum width=3cm, minimum height=1cm, text centered, draw=black, fill=blue!30]
\tikzstyle{process} = [rectangle, minimum width=3cm, minimum height=1cm, text centered, draw=black, fill=orange!30]
\tikzstyle{decision} = [diamond, minimum width=3cm, minimum height=1cm, text centered, draw=black, fill=green!30]
\tikzstyle{arrow} = [thick,->,>=stealth]

\colorlet{punct}{red!60!black}
\definecolor{background}{HTML}{ffffff }
\definecolor{delim}{RGB}{20,105,176}
\colorlet{numb}{magenta!60!black}

\definecolor{light-gray}{gray}{0.95}
\definecolor{darkgray}{rgb}{0.4, 0.4, 0.4}
\definecolor{editorGray}{rgb}{0.95, 0.95, 0.95}
\definecolor{editorOcher}{rgb}{1, 0.5, 0} 
\definecolor{editorGreen}{rgb}{0, 0.5, 0} 
\definecolor{orange}{rgb}{1,0.45,0.13}      
\definecolor{olive}{rgb}{0.17,0.59,0.20}
\definecolor{brown}{rgb}{0.69,0.31,0.31}
\definecolor{purple}{rgb}{0.38,0.18,0.81}
\definecolor{lightblue}{rgb}{0.1,0.57,0.7}
\definecolor{lightred}{rgb}{1,0.4,0.5}
\usepackage{upquote}
\usepackage{listings}

\definecolor{pblue}{rgb}{0.13,0.13,1}
\definecolor{pgreen}{rgb}{0,0.5,0}
\definecolor{pred}{rgb}{0.9,0,0}
\definecolor{pgrey}{rgb}{0.46,0.45,0.48}

\begin{document}

\title{Self-Supervised Graph Transformer for
\\
Deepfake Detection }

\author{Aminollah Khormali,~\IEEEmembership{Student Member,~IEEE,}
        Jiann-Shiun Yuan,~\IEEEmembership{Senior Member,~IEEE}
\IEEEcompsocitemizethanks{
\IEEEcompsocthanksitem A. Khormali and J. S. Yuan are with the Department of Electrical and Computer Engineering, University of Central Florida, Orlando, Fl, USA. 
}
}

\IEEEoverridecommandlockouts
\makeatletter\def\@IEEEpubidpullup{9\baselineskip}\makeatother

\maketitle

\begin{abstract}
Deepfake detection methods have shown promising results in recognizing forgeries within a given dataset, where training and testing take place on the in-distribution dataset. However, their performance deteriorates significantly when presented with unseen samples. As a result, a reliable deepfake detection system must remain impartial to forgery types, appearance, and quality for guaranteed generalizable detection performance. Despite various attempts to enhance cross-dataset generalization, the problem remains challenging, particularly when testing against common post-processing perturbations, such as video compression or blur. Hence, this study introduces a deepfake detection framework, leveraging a self-supervised pre-training model that delivers exceptional generalization ability, withstanding common corruptions and enabling feature explainability. The framework comprises three key components: a feature extractor based on vision Transformer architecture that is pre-trained via self-supervised contrastive learning methodology, a graph convolution network coupled with a Transformer discriminator, and a graph Transformer relevancy map that provides a better understanding of manipulated regions and further explains the model’s decision. To assess the effectiveness of the proposed framework, several challenging experiments are conducted, including in-data distribution performance, cross-dataset \& cross-manipulation generalization, and robustness against common post-production perturbations. The results achieved demonstrate the remarkable effectiveness of the proposed deepfake detection framework, surpassing the current state-of-the-art approaches.
\end{abstract}


\section{Introduction}\label{sec:Introduction}
\input{introduction}
\section{Related Work}\label{sec:RelatedWork}
\input{related}
\section{Methodology}\label{sec:Methodology}
\input{methodology}

\section{Evaluation Settings}\label{sec:EvaluationSettings}
\input{evaluation}

\section{Results \& Discussion}\label{sec:Results}
\input{results}

\section{Conclusion}\label{sec:Conclusion}

\input{conclusion}

\bibliographystyle{IEEEtran}
\bibliography{ref.bib}

\end{document}

%% file: introduction.tex
The rise of Artificial Intelligence (AI) empowered face manipulation/generation technologies have enabled individuals' expressions or appearances to be realistically altered with minimal expert knowledge \cite{dolhansky2019deepfake, jiang2020deeperforensics, li2020advancing, Celeb_DF_cvpr20, rossler2019faceforensics++, zi2020wilddeepfake}. This technology poses a severe and large-scale societal threat, as it facilitates the creation and dissemination of malicious content, such as digital kidnapping, ransomware, and other forms of criminal activity \cite{chen2021defakehop, khormali2021add} that are among the most insidious forms of misinformation \cite{tran2021high}. Consequently, the development of reliable deepfake detection methods has become an urgent need and has garnered significant attention in recent years. Although several efforts have been made to defend against the growing threat of forged digital media, the effectiveness of these approaches have primarily been limited to in-dataset settings \cite{agarwal2020detecting, haliassos2021lips, li2020face, liu2021spatial, masi2020two, qian2020thinking, rossler2019faceforensics++, zhao2021multi, khormali2022dfdt}. Therefore, the need to develop more robust and effective deepfake detection methods that can operate in real-world scenarios irrespective of the forgery type, appearance, or quality has become increasingly pressing.

In the domain of deepfake detection, previous investigations have predominantly relied upon low-level texture features to capture common artifacts that are inherent to the generation process. Nonetheless, such approaches are susceptible to severe performance degradation when applied to novel types of forgeries, rendering the detection of real-world deepfakes a challenging task, especially when the differences between genuine and manipulated media are nuanced. As a result, an effective deepfake detection model ought to be impartial towards forgery type, appearance, and quality, in order to ensure applicability in real-world scenarios. To this end, several techniques have been proposed to enhance the performance and generalization of deepfake detectors, such as targeting the blending boundary between the background and the altered face \cite{li2020face}, utilizing 3D decomposition \cite{zhu2021face}, truncating classifiers \cite{chai2020makes}, amplifying multi-band frequencies \cite{masi2020two}, and augmenting data \cite{wang2020cnn}. Despite their effectiveness in cross-data evaluations, low-level texture cues may be vulnerable to degradation through standard post-processing procedures, such as video compression \cite{haliassos2021lips}. Consequently, it is imperative to develop novel high-level features that are invariant to image manipulations and exhibit resilience to post-processing manipulations to enhance the current state-of-the-art in deepfake detection. 

Moreover, The widely-used convolutional neural network and Transformer treat the image as a grid or sequence structure, which is not flexible to capture irregular and complex facial landmarks. Instead of the regular grid or sequence representation, this work processes the deepfake image in a more flexible way. Human facial images can be viewed as a composition of parts, such as eyes, ears, nose, etc. These facial regions naturally form a graph structure. This graph representation of deepfakes offers a better understanding of local pixels and their interconnection with other parts.  Moreover, the graph is a generalized data structure that grid and sequence can be viewed as a special case of the graph. Viewing an image as a graph is more flexible and effective for visual perception.

Given the limitations of existing methods, building a detection model that capitalizes on high-level visual representations, which tend to be more resilient to post-processing perturbations, could potentially mitigate the dependence on dataset-specific patterns, leading to improved recognition performance across various forgery types. In light of this, this study first proposes a self-supervised pre-training approach for deepfake facial representation learning. This approach utilizes a contrastive learning framework based on masked image modeling to extract high-level visual representations of facial landmarks that are invariant to variations in lighting, pose, and other factors that are irrelevant to the identity of the object or scene. The contrastive learning method maximizes the agreement between two different augmentations of the same image using a contrastive loss in the latent space \cite{chen2020simple, he2020momentum, tschannen2020self}. This means that the learned representation is not biased towards any particular class or label, but rather captures the underlying structure of the input data.  

Moreover, the present work offers an innovative approach that integrates the unique characteristics of Graph Convolutional Networks (GCNs) and Transformer architectures to capture complex dependencies between distinct regions of an image and acquire more informative representations for deepfake detection. In the context of deepfake detection, the input data can be projected as a graph, wherein nodes denote various regions of the image, such as eyes or ears, and edges represent the association between these regions. This is particularly useful in detecting deepfakes, where only certain regions of the image may be manipulated to create realistic but false identities. The utilization of GCNs allows for modeling local relationships between nodes in the input graph, providing the ability to learn complex features that capture the spatial correlations between different facial landmarks. In contrast, Transformers exhibit exceptional efficacy in encoding long-range correlations and global interdependencies between pixels, rendering them especially appropriate for deepfake detection tasks. Given that adversarial entities can manipulate multiple regions of an image concurrently, the ability to model such complex relationships is crucial for reliable detection. Finally, this work introduces a graph Transformer relevancy map that contributes valuable insights into the model's explainability. By generating a saliency map that highlights the importance of individual regions towards the output class label, this component facilitates a more fine-grained analysis of the model's decision-making process and allows for a better understanding of the specific features and regions of the image that are crucial for deepfake detection. \autoref{fig:SSLGT} shows the general framework of the proposed deepfake detection model using a self-supervised contrastive learning approach and graph Transformer architecture.

\BfPara{Contributions}
Together, these contributions offer a promising solution to the urgent need for robust and effective deepfake detection methods that can operate irrespective of the forgery types, appearances, or qualities in real-world scenarios. The proposed framework's use of high-level visual representations that are invariant to post-processing perturbations, combined with the ability to capture complex interdependencies between regions of an image, provides a comprehensive solution that outperforms existing methods. Below are the key contributions of this study:

\begin{itemize}

    \item The present study introduces a self-supervised pre-training approach based on contrastive learning to extract more high-level visual representations that are invariant to variations in compression, blur, and other factors irrelevant to the subject's identity. This approach significantly enhances the generalizability of deepfake detection models, thereby reducing their dependence on dataset-specific patterns. 

    \item Furthermore, this paper proposes an innovative deepfake detection approach by leveraging the expressive power of graph convolutional networks and Transformers to capture intricate interdependencies among different regions of an image and acquire more informative representations. Unlike traditional convolutional neural networks, which are limited in their ability to model non-local relationships between pixels, the proposed graph Transformer can encode both local and global dependencies, rendering them particularly suited to deepfake detection tasks.
    
    \item  By providing detailed and granular insights into the underlying reasoning behind the model's predictions, the proposed graph Transformer relevancy map offers a more thorough and rigorous understanding of the complex interdependencies and salient features that drive the detection process, while suppressing irrelevant or redundant information. Thus, facilitating a rigorous and accurate examination of deepfakes, contributing to enhanced detection performance and increased reliability.
       
    \item The proposed framework's efficacy and generalizability were rigorously evaluated via a comprehensive set of experiments, encompassing a diverse range of challenging scenarios spanning both in-distribution and out-of-distribution settings. The experimental results unequivocally establish the framework's superiority, with exceptional in-dataset detection accuracy being achieved. Moreover, the proposed self-supervised pre-training feature extractor constitutes a significant contribution to the field, having markedly improved the framework's ability to generalize across multiple datasets while simultaneously enhancing its resilience to post-processing perturbations, such as compression and blur \etc
    
\end{itemize}

\BfPara{Organization} The rest of the paper is structured as follows. 
In \textsection\ref{sec:RelatedWork}, a review of recent research at the intersection of deepfake detection and self-supervised contrastive learning is provided. The proposed self-supervised graph Transformer deepfake detection approach, including its key components and the rationale for their selection, is outlined in \textsection\ref{sec:Methodology}.  The evaluation settings, including datasets, implementation specifics, and evaluation metrics, are described in~\textsection\ref{sec:EvaluationSettings}. In~\textsection\ref{sec:Results}, the obtained deepfake detection results and their implications are presented and discussed in comparison with existing methods. Finally, in~\textsection\ref{sec:Conclusion}, concluding remarks are drawn to summarize the contributions of this study and outline avenues for future research.

%% file: related.tex
\begin{figure*}[t]
    \centering
    \includegraphics[width=0.95\textwidth]{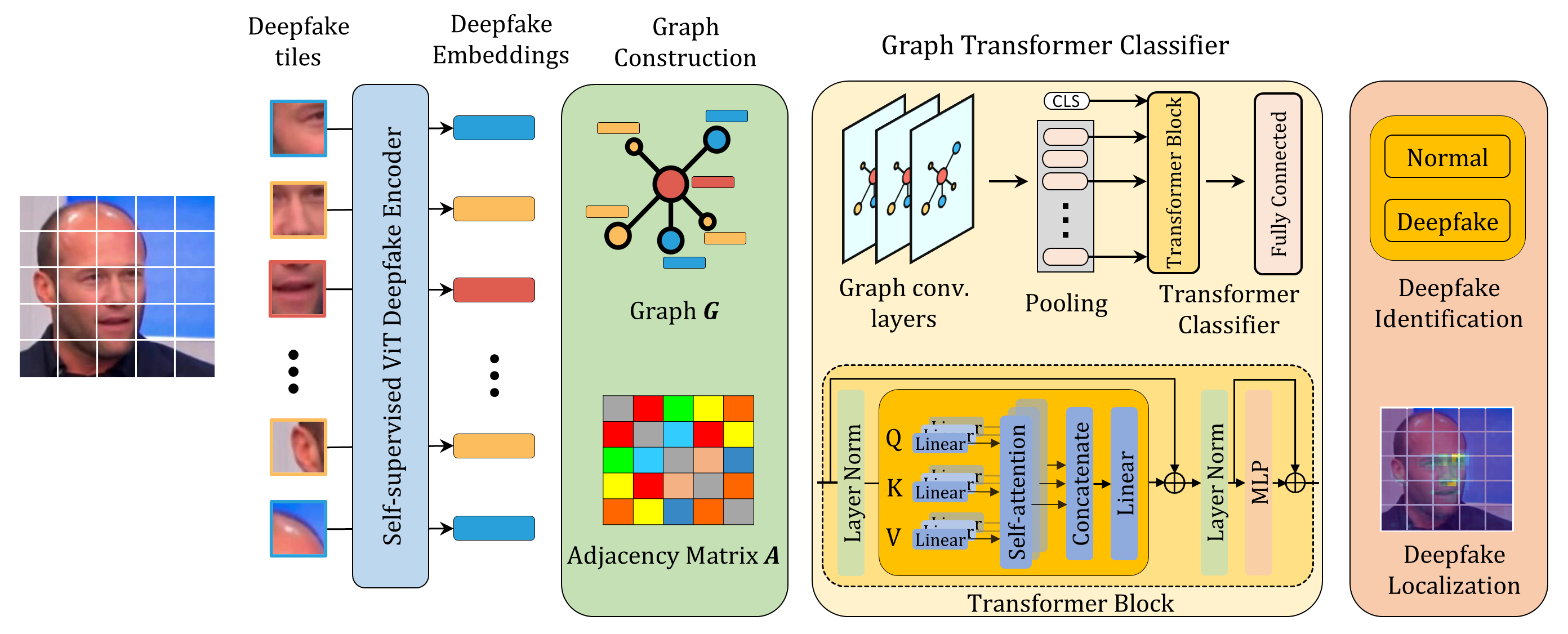}
    \caption{
    The general framework of the proposed deepfake detection model using a self-supervised contrastive learning approach and graph Transformer architecture. In this approach, each input image is partitioned into smaller patches, where each patch is considered as a node in an undirected graph. A feature extractor model based on vision transformer architecture, which is pre-trained based on contrastive learning methodology, is then applied to each patch to obtain its visual features. These nodes and associated feature vectors are restructured in the form of a graph based on the adjacency matrix. The constructed graph is subsequently fed into a graph convolutional network, which is followed by a Transformer network that functions as a deepfake discriminator, with the purpose of forecasting image class labels. Finally, the Transformer relevancy map is utilized to identify potentially manipulated facial regions. The suggested deepfake detection mechanism capitalizes on the merits of both high-level visual characteristics and the interconnectivity of local and global regions through the pre-trained feature extractor and graph Transformer classifier, respectively.  
    }
    \label{fig:SSLGT}
\end{figure*}

\subsection{Deepfake Detection}
Recent years have witnessed a growing body of research aimed at addressing the pressing challenges posed by deepfakes and developing robust detection models capable of thwarting their nefarious effects on society. A variety of techniques have been proposed in the literature to this end, with a rich diversity of approaches being explored \cite{agarwal2020detecting, haliassos2021lips, khormali2021add, khormali2022dfdt, li2020face, liu2021spatial, wang2022m2tr, zhao2021multi}.

One popular strategy is to leverage the implicit visual artifacts of deepfakes' generation process to construct a reliable detection framework \cite{li2020face, matern2019exploiting}. These approaches capitalize on the inconsistencies and discrepancies that inevitably arise during the creation of deepfakes, such as blending artifacts or anomalies in the frequency spectrum. Examples of such approaches include Patch-wise Consistency Learning \cite{zhao2021learning} and Face X-ray \cite{li2020face}, which exploit image blending inconsistency traces between the manipulated face region and the background. These approaches exemplify the diverse range of methods used to detect deepfakes, with some leveraging specific visual artifacts while many mainstream deepfake detection methodologies rely heavily on the versatility of convolutional neural networks. For instance, \cite{bonettini2021video, khormali2021add, marra2019gans, rossler2019faceforensics++} rely on the exceptional ability of convolutional neural networks to capture complex spatial features and relationships as the primary building block of their proposed deepfake detection framework.

Additionally, recent research has explored alternative avenues, such as analyzing the temporal dynamics of videos or using frequency spectrum analysis to detect deepfakes based on unique spectral signatures. For instance, the temporal dynamics of videos have been explored as a means to recongnize fake videos \cite{dang2020detection, sabir2019recurrent, zhao2023istvt, yin2023dynamic}. This approach involves examining the directional changes in the video's temporal characteristics, thereby detecting any inconsistencies or abnormal patterns indicative of deepfakes. Furthermore, frequency spectrum analysis is gaining significant traction in research interest \cite{durall2020watch, frank2020leveraging, li2021frequency, liu2021spatial, luo2021generalizing, masi2020two, qian2020thinking}. This technique capitalizes on leveraging the identification of spectral signatures exhibited by deepfakes, such as the existence of high-frequency noise or low-pass filtering artifacts, enabling the detection of manipulated content.


In light of recent advancements in deepfake generation techniques that have reduced the visual anomalies, the detection methods have shifted focus towards more sophisticated approaches such as attention mechanisms \cite{dang2020detection, wang2021representative, zhao2021multi, khormali2021add, yang2023masked} and vision Transformers \cite{khormali2022dfdt, wang2022m2tr, wang2022adt}. For example, Wang et al. \cite{wang2022m2tr} proposed a multi-regional attention mechanism to enhance deepfake detection performance. Additionally, vision Transformers have been employed to establish an end-to-end deepfake detection framework \cite{khormali2022dfdt, wang2022lisiam}. Furthermore, recent research has utilized pre-trained networks and developed Lipforensics \cite{haliassos2021lips} for analyzing lip prints in lip-reading tasks. This work highlights that while a robust pre-trained lip feature extractor could have far-reaching implications, acquiring a large-scale and well-annotated dataset could be a daunting task. Given the potential limitations of relying on annotated data, this study proposes a self-supervised vision Transformer for pretraining the feature extractor without the need for annotation or labels. The resulting framework is scalable and achieves good performance.

\subsection{Self-Supervised Contrastive Learning}
A meaningful visual representation of a given input image could be learned through self-supervised contrastive learning methods without relying on labeled data \cite{oord2018representation}. Recently, self-supervised contrastive learning approaches have gained tremendous attention in building pre-trained models that can be used for fine-tuning different downstream tasks. Contrastive learning maximizes the similarity between positive pairs while repelling features of negative pairs \cite{chen2020simple}. Although contrastive learning approaches have been utilized for dense visual representation learning tasks \cite{o2020unsupervised, wang2021dense}, recent studies suggest that removing negative pairs yields better performances \cite{grill2020bootstrap, caron2020unsupervised, caron2021emerging, chen2021exploring}. While Random augmentations have been widely used to generate diverse views of the same image \cite{chen2020simple, chen2021empirical, fung2021deepfakeucl}, this work, inspired by  \cite{bao2022beit}, proposes a masked image modeling approach to extract more robust semantic features through a contrastive learning methodology. In contrast to previous work that only focused on mouth regions for lip-forensics~\cite{zhao2022self}, this work leverages the entire facial region to improve the detection performance. The proposed masked image modeling approach was utilized to train a pre-trained model, with a vision Transformer serving as its backbone. Through this approach, complex semantic features from input images were captured by the model, resulting in improved deepfake detection performance.

%% file: methodology.tex

The proposed deepfake detection method entails four fundamental components, as illustrated in \autoref{fig:SSLGT}, which are deepfake graph construction, deepfake encoder based on vision Transformer architecture that is pre-trained using a self-supervised contrastive learning approach, graph Transformer classifier, and graph Transformer relevancy map. This section is dedicated to elaborating on each of these components in a more comprehensive manner.

\subsection{Deepfake Graph Representation}
The irregularity of facial regions' shape constitutes a major obstacle in employing conventional techniques, such as grid-based architectures in CNNs or sequence-based structures in Transformers, to process facial images. Such approaches are characterized by redundancy and inflexibility, rendering them incapable of efficiently addressing the complex structural nuances of facial regions. As a consequence, there is a need for more sophisticated and adaptive techniques that can account for the varying geometries and topologies of these regions. In this regard, the present study advocates the adoption of graph representation of deepfakes as a highly promising method to tackle the issue of deepfake detection, owing to its capacity to account for the varying topologies and geometries of the facial regions. By leveraging this approach to represent deepfake images as graphs, it becomes possible to capture the underlying structural and relational features of key facial landmarks, such as the eyes, lips, and ears. This capability, in turn, enables the identification of potentially manipulated regions and their relationship within the image, thereby capturing more complex features for enhanced and robust deepfake detection.

The construction of an undirected graph from an image input is a pivotal step in the proposed deepfake detection methodology. In essence, the process of representing a given input image $I$ as a graph $G = (V, E)$ involves partitioning it into $N$ patches, where $V = \left\{ v_{1}, v_{2}, \dots, v_{N} \right\}$ signifies the nodes of the graph and $E$ denotes its corresponding edges. This entails representing feature embeddings extracted from image patches as graph nodes, denoted by $v_i \in \mathbb{R}^D$, where $D$ is the dimensionality of the embedding vector. The adjacency relationship among these patches constitutes the graph edges $E$. Notably, edges within the graph are established by means of a spatial proximity criterion that determines the association of each patch with its neighboring counterparts, allowing for at least one, and up to $K$ connections between $K$ nearest neighboring patches. To encode the adjacency relationship between patches, an adjacency matrix, $A = [A_{ij}]$, is formulated, is formulated, where $A_{ij} = A_{ji} = 1$ if the nodes $v_{i}$ and $v_{j}$ are in the neighborhoods, and $A_{ij} = A_{ji} = 0$ otherwise, that is if $(v_i, v_j) \notin E$. By modeling deepfake images in this way, the structural relationships and dependencies among the various facial components can be captured and analyzed in a more comprehensive manner. Furthermore, the efficacy of the proposed methodology depends crucially on the feature vector's ability to provide a potent and resilient representation of the node in question. To achieve this, the present study utilizes self-supervised contrastive learning and a vision Transformer structure to extract high-level visual representations of deepfake images. This enables the detection system to capture the salient attributes of the underlying image regions, resulting in more informative and discriminative features. As high-level features are less vulnerable to manipulations, this approach maximizes the model's potential to generalize to new data and perform well on downstream tasks.

\begin{figure*}[t]
    \centering
    \includegraphics[width=0.8\textwidth]{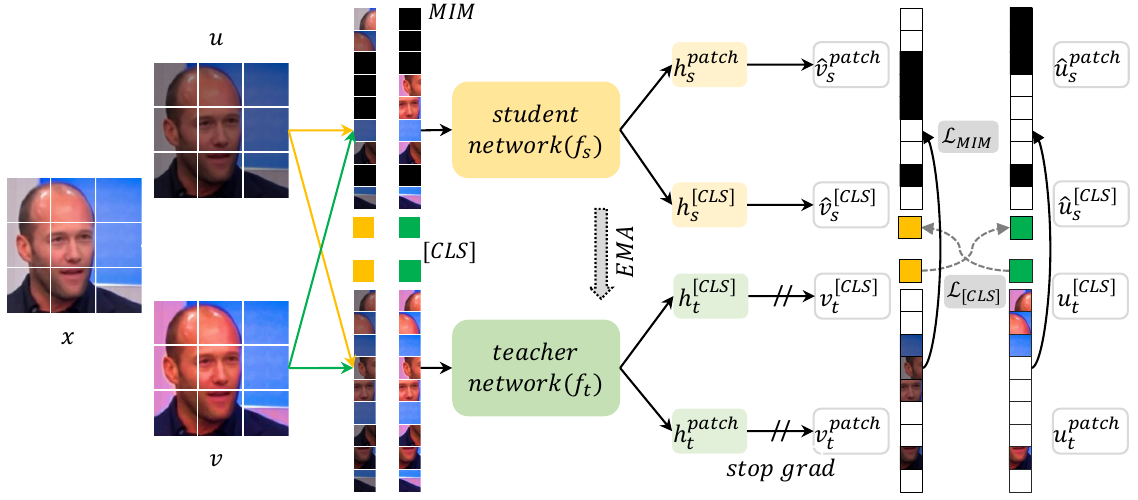}
    \caption{General framework of the vision Transformer-based feature extractor trained through self-supervised contrastive learning. Given a deepfake image, $x$, two different views, $u$ and $v$, are passed through student and teacher networks. The student network tries to reconstruct masked tokens with the supervision of teacher networks' output by minimizing self-distillation between cross-view $[\mathrm{CLS}]$ tokens and self-distillation between in-view patch tokens.}
    \label{fig:SSL}
\end{figure*}

\subsection{Self-Supervised Contrastive Learning} \label{sec:SSL}

Existing deepfake detection methodologies heavily rely on weakly supervised learning approaches for visual feature extraction. While such approaches exhibit high detection performance on in-distribution samples, their ability to generalize to unseen samples is limited. This generalizability issue poses a significant challenge to real-world applications where the model must be robust to unseen scenarios. To overcome this issue, this study proposes the utilization of self-supervised contrastive learning for robust visual feature extraction without the need for explicit labels. Specifically, the proposed approach employs masked image modeling, which applies a random mask to an input image with $N$ token sequences and a prediction ratio of $r$, to train a vision Transformer for generating patch representations. The model maximizes the concordance between two distinct views of the same image through contrastive loss in a latent space. This approach is inspired by recent studies on self-supervised contrastive learning, including \cite{chen2020simple}, and builds on the successful application of masked image modeling in \cite{bao2022beit, zhou2022image}. Figure \ref{fig:SSL} illustrates the proposed self-supervised contrastive learning framework for deepfake representation learning. The resulting visual features are expected to be more informative and discriminative, improving the model's ability to generalize to unseen data and achieve high performance on downstream tasks.

The proposed training methodology commences with the random selection of an image $x$ from the training set, followed by the application of random augmentations to generate two distorted views of the same image, denoted as $\boldsymbol{u}$ and $\boldsymbol{v}$. To enhance robustness, blockwise masking \cite{bao2022beit} is then applied to produce masked views of the augmented images, denoted as $\hat{\boldsymbol{u}}$ and $\hat{\boldsymbol{v}}$. Subsequently, the masked views undergo processing by a student-teacher network to yield predictive categorical distributions. Both the student and teacher networks share an identical architectural configuration and are parameterized by $\boldsymbol{\theta}$ and $\boldsymbol{\theta}^{\prime}$, respectively. This framework enables the self-supervised training of the student network, which endeavors to reconstruct the masked tokens by leveraging the guidance of the teacher network's output.

In self-supervised frameworks, it is customary for the student network to acquire distilled knowledge from the teacher network by minimizing the cross-entropy loss \cite{he2020momentum, grill2020bootstrap, caron2021emerging}. However, in the present study, the deepfake images' visual representations are acquired through the minimization of two concurrent loss functions, namely: 1) $\mathcal{L}{[\mathrm{CLS}]}$, which represents the self-distillation between cross-view $[\mathrm{CLS}]$ tokens, and 2) $\mathcal{L}{[\mathrm{MIM}]}$, which represents the self-distillation between in-view patch tokens. Specifically, the self-distillation process regarding the $[\mathrm{CLS}]$ tokens of cross-view images, denoted by $\boldsymbol{u}$ and $\boldsymbol{v}$, can be formulated as a symmetric loss, as shown in Equation \ref{eq:losscls}.

\begin{equation}\label{eq:losscls}
\begin{aligned}
\mathcal{L}_{[\mathrm{CLS}]} &=\frac{1}{2} \left(-P_{\boldsymbol{\theta}^{\prime}}^{[\mathrm{CLS}]}(\boldsymbol{v})^{\mathrm{T}} \log P_{\boldsymbol{\theta}}^{[\mathrm{CLS}]}(\boldsymbol{u}) \right) + \\
&\frac{1}{2} \left(-P_{\boldsymbol{\theta}^{\prime}}^{[\mathrm{CLS}]}(\boldsymbol{u})^{\mathrm{T}} \log P_{\boldsymbol{\theta}}^{[\mathrm{CLS}]}(\boldsymbol{v}) \right)
\end{aligned}
\end{equation}

where  ${\boldsymbol{u}}_{s}^{\mathrm{CLS}}=P_{\boldsymbol{\theta}}^{\mathrm{CLS}}({\boldsymbol{u}})$, ${\boldsymbol{v}}_{s}^{\mathrm{CLS}}=P_{\boldsymbol{\theta}}^{\mathrm{CLS}}({\boldsymbol{v}})$, $\boldsymbol{u}_{t}^{\mathrm{CLS}}=P_{\boldsymbol{\theta}^{\prime}}^{\mathrm{CLS}}(\boldsymbol{u})$, and $\boldsymbol{v}_{t}^{\mathrm{CLS}}=P_{\boldsymbol{\theta}^{\prime}}^{\mathrm{CLS}}(\boldsymbol{v})$ are student and teacher network outputs for cross-view images, \ie $\boldsymbol{u}$ and $\boldsymbol{v}$, of [CLS] token.

The student and teacher network outputs for the masked view with $m_i$ masking ratio for  $\hat{\boldsymbol{u}}$,  $\hat{\boldsymbol{v}}$ and non-masked view $\boldsymbol{u}$, $\boldsymbol{v}$ projections of patch tokens are claculated as $\hat{\boldsymbol{u}}_{s}^{\mathrm{patch}}=P_{\boldsymbol{\theta}}^{\mathrm{patch}}(\hat{\boldsymbol{u}})$,  $\hat{\boldsymbol{v}}_{s}^{\mathrm{patch}}=P_{\boldsymbol{\theta}}^{\mathrm{patch}}(\hat{\boldsymbol{v}})$ and  $\boldsymbol{u}_{t}^{\mathrm{patch}}=P_{\boldsymbol{\theta}^{\prime}}^{\mathrm{patch}}(\boldsymbol{u})$,  $\boldsymbol{v}_{t}^{\mathrm{patch}}=P_{\boldsymbol{\theta}^{\prime}}^{\mathrm{patch}}(\boldsymbol{v})$, respectively. Similarly, the self-distillation between in-view patch tokens can be formulated as a symmetric cross-entropy loss as \autoref{eq:lossmim}.
\begin{equation}\label{eq:lossmim}
\begin{aligned}
\mathcal{L}_{\mathrm{MIM}}= &=\frac{1}{2} \left(-\sum_{i=1}^{N} m_{i} \cdot P_{\boldsymbol{\theta}^{\prime}}^{\mathrm{patch}}\left(\boldsymbol{u}_{i}\right)^{\mathrm{T}} \log P_{\boldsymbol{\theta}}^{\mathrm{patch}}\left(\hat{\boldsymbol{u}}_{i}\right) \right)+ \\
&\frac{1}{2} \left( -\sum_{i=1}^{N} m_{i} \cdot P_{\boldsymbol{\theta}^{\prime}}^{\mathrm{patch}}\left(\boldsymbol{v}_{i}\right)^{\mathrm{T}} \log P_{\boldsymbol{\theta}}^{\mathrm{patch}}\left(\hat{\boldsymbol{v}}_{i}\right)\right)
\end{aligned}
\end{equation}

The overall loss function is expressed as a combination of two individual loss functions, $\mathcal{L}_{[\mathrm{CLS}]}$ and $\mathcal{L}_{\mathrm{MIM}}$, as presented in \autoref{eq:loss2}. The loss function $\mathcal{L}_{[\mathrm{CLS}]}$ involves self-distillation between cross-view $[\mathrm{CLS}]$ tokens, whereas $\mathcal{L}_{\mathrm{MIM}}$ is related to self-distillation among in-view patch tokens. 
The proposed approach leverages the collective information obtained from both loss functions to effectively learn the deepfake image representations.

\begin{equation}\label{eq:loss2}
\mathcal{L}= \mathcal{L}_{[\mathrm{CLS}]}+ \mathcal{L}_{\mathrm{MIM}}
\end{equation}

\subsection{Graph Transformer Classifier}

The methodology employed in this study involves the use of a vision Transformer as a feature extractor, which has been trained using a self-supervised contrastive learning approach, as previously mentioned in \autoref{sec:SSL}. This approach is effective in computing the visual representation of each node in the graph. An image representation matrix $\mathcal{F} \in \mathbb{R}^{N \times D}$ is constructed by subjecting all $N$ nodes within the image to the aforementioned pre-trained ViT feature extractor. Specifically, each node $v_{i}$ is associated with a feature vector $f_{i} \in \mathbb{R}^{D}$, where $D$ denotes the dimensionality of the feature vector. The resultant matrix is defined as $F = \left\{ f_{1}, f_{2},\dots, f_{N} \right\}$. It is noteworthy that the combination of $F$ with the adjacency matrix $\mathcal{A}$ facilitates the construction of a graph representation of deepfake images. The resulting graph-structured data can be analyzed using a multi-layer graph convolutional network $f(\mathcal{F}, \mathcal{A})$ \cite{kipf2016semi}, which utilizes a layer-wise propagation rule as specified in \autoref{eq:gclayer}.

\begin{equation}\label{eq:gclayer}
\begin{array}{l}
H_{l+1}=R e L U\left(\hat{A} H_{l} W_{l}\right), \quad l=1,2, . ., L \\
\hat{A}=\tilde{D}^{-\frac{1}{2}} \tilde{A} \tilde{D}^{-\frac{1}{2}}
\end{array}
\end{equation}

here, $L$ represents the total number of graph convolution layers, and the activation matrix of the $l^{th}$ graph convolutional layer is denoted as $H_l$. The activation matrix $H_l$ captures the learned representations at each layer, with the initialization of $H_0$ as the input feature matrix $F$. The weight matrix $W_l$ is specific to each layer and is learned during training. Moreover, the graph's adjacency matrix with added self-connections to each node is represented by $\tilde{A}$ and is defined as the sum of the adjacency matrix $A$ and the identity matrix $I$. A diagonal matrix $\tilde{D}$ is used to normalize the adjacency matrix, where $\tilde{D}_{ii}$ is defined as the sum of the entries in row $i$ of the normalized adjacency matrix $\tilde{A}$. The resulting matrix $\hat{A}$ is the symmetric normalized adjacency matrix of $A$.

Although graph convolutional layers have been successful in learning node-level features and have been widely used in graph neural networks, they have limitations in learning hierarchical visual features that capture the global context of the graph. In contrast, the attention mechanism in Transformer models has shown remarkable success in natural language processing tasks by capturing the long-range dependencies between tokens and their relative importance for the final prediction \cite{vaswani2017attention}. This ability is particularly important in graph-structured deepfake images, where the relationships between nodes can be complex and far-reaching. Therefore, the present work proposes to leverage attention mechanisms to analyze graph-structured data by treating the feature nodes as tokens in a sequence, and adjacency matrices as positional encodings to preserve their spatial relationship. By synergizing the attention mechanism with graph convolutional layers, the proposed model proficiently assimilates both localized and global features, and adeptly captures intricate inter-nodal relationships within the graph. Notably, the transformation of graph space data into Transformer space is enacted via a Transformer layer, as specified in \autoref{eq:graphattention1} to \autoref{eq:graphattention4}.


\begin{equation}\label{eq:graphattention1}
\mathbf{z}_{0} =\left[x_{\mathrm{class}} ; h^{(1)} ; h^{(2)} ; \ldots ; h^{(N)}\right], \quad h^{(i)} \in H
\end{equation}
\begin{equation}
\mathbf{z}_{\ell}^{\prime}=\operatorname{MSA}\left(\mathrm{LN}\left(\mathbf{z}_{\ell-1}\right)\right)+\mathbf{z}_{\ell-1}, \quad \quad \ell=1 \ldots L
\end{equation}
\begin{equation}
\mathbf{z}_{\ell}=\operatorname{MLP}\left(\operatorname{LN}\left(\mathbf{z}_{\ell}^{\prime}\right)\right)+\mathbf{z}_{\ell}^{\prime}, \quad \ell=1 \ldots L
\end{equation}
\begin{equation}\label{eq:graphattention4}
\mathbf{y}=\mathrm{LN}\left(\mathbf{z}_{L}^{0}\right)
\end{equation}


Here, the model architecture entails a multihead self-attention mechanism $\operatorname{MSA}$, featuring $k$ Self-Attention (SA) heads, as delineated in Equation \ref{eq:attention} and \ref{eq:MSA}, respectively. Furthermore, $\operatorname{MLP}$ represents a Multilayer Perceptron, and Layer Norm is indicated by $\mathrm{LN}$. Here, $L$ corresponds to the number of Transformer blocks, and $\mathbf{y}$ signifies the class label \cite{dosovitskiy2020image}. The Transformer model utilizes the graph feature embeddings, as outlined in Equation \ref{eq:gclayer}, to facilitate self-attention operation $\text{SA}(Q, K, V)$, based on Transformer architecture as presented in Equation \ref{eq:attention}.


\begin{equation}\label{eq:MSA}
\begin{aligned}
\text { MSA }(Q, K, V) &=\text { Concat }\left(\operatorname{SA}_{1}, \ldots, \text { SA}_{\mathrm{k}}\right) W^{O} \\
\text { where SA}_i &=\text { SA }\left(Q W_{i}^{Q}, K W_{i}^{K}, V W_{i}^{V}\right)
\end{aligned}
\end{equation}

\begin{equation}\label{eq:attention}
\text { SA }(Q, K, V)=\operatorname{softmax}\left(\frac{Q K^{T}}{\sqrt{d_{k}}}\right) V
\end{equation}

where, $N$ represents the number of patches, $D$ is the dimension of patch embeddings, and $\text{SA}(Q, K, V)$ denotes the pairwise similarity of two nodes based on their corresponding Query $(Q_{in})$, Key $(K)$, and Values $(V)$. Notably, the projections are parameter matrices $W_{i}^{Q} \in \mathbb{R}^{D \times d_{k}}, W_{i}^{K} \in \mathbb{R}^{D \times d_{k}}, W_{i}^{V} \in \mathbb{R}^{D \times d_{v}}$ and $W^{O} \in \mathbb{R}^{kd_{v} \times D}$ \cite{vaswani2017attention, dosovitskiy2020image}.

Finally, it is imperative to acknowledge that the conventional self-attention mechanism computes attention scores for all possible pairs of nodes, leading to an exponential increase in memory and time complexity as the number of nodes increases, with a complexity of $O(n^2)$. Thus, a technique that reduces the number of nodes while retaining local information is indispensable. In this investigation, a learnable pooling layer, referred to as the min-cut pooling layer \cite{bianchi2020spectral}, has been employed on the output of the final graph convolutional layer to effectively reduce the number of nodes.

\begin{figure}[t]
    \centering
    \includegraphics[width=0.5\textwidth]{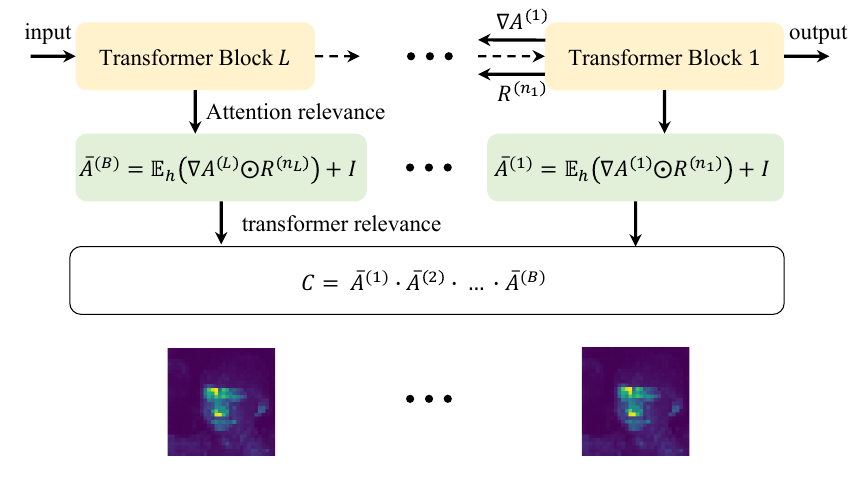}
    \caption{ Relevancy maps for the Transformer, $C$, are generated via the integration of attention maps, gradients, and relevance scores, which are propagated throughout the network. These maps are subsequently converted into graph class activation maps through reverse pooling, as elucidated in \autoref{eq:cam}. 
    }
    \label{fig:graphcam}
\end{figure}

\subsection{Graph Transformer Relevancy Map}
To pinpoint areas within a given image that are likely to have been manipulated, and which exhibit a significant correlation with deepfake class labels, the present study builds upon previous research in this field \cite{chefer2021transformer} and presents a graph Transformer relevancy map. This innovative approach involves computing the class activation maps from the output class and then propagating them to the inner graph space through a propagation procedure. The current study shows that the attention relevancy of the $\ell^{th}$ Transformer block  $\bar{\mathbf{A}}^{(\ell)}$ can be obtained by representing the attention map of its Transformer block as $A^{\ell}$, which facilitates the calculation of the layer relevance score $R^{n_\ell}$ and gradient $\nabla A^{(\ell)}$ values pertaining to class $t$. It should be noted that ${n_\ell}$ corresponds to the layer that aligns with the $softmax$ operation in the $\ell^{th}$ Transformer block. Finally, as depicted in \autoref{fig:graphcam} the Transformer's final relevancy maps are represented by a weighted attention relevance, as described by \autoref{eq:cam}. 

\begin{equation}\label{eq:cam}
\begin{aligned}
\bar{\mathbf{A}}^{(\ell)} &=\mathbb{E}_{h}\left(\nabla \mathbf{A}^{(\ell)} \odot R^{\left(n_{l}\right)}\right) + I\\
\mathbf{C_{t}} &=\bar{\mathbf{A}}^{(1)} \cdot \bar{\mathbf{A}}^{(2)} \cdot \ldots \cdot \bar{\mathbf{A}}^{(L)}
\end{aligned}
\end{equation}


Here, $\mathbb{E}_{h}$ signifies the mean value calculated across the dimension of the self-attention heads of the Transformer model, whereas $\odot$ denotes the Hadamard product. To prevent self-inhibition for each node, the identity matrix $I$ is added. In the final step, the Transformer relevance map $C_{t}$ is mapped to each node in the graph, based on the min-cut dense learned assignments \cite{bianchi2020spectral}. This min-cut approach ensures that the most relevant nodes are assigned the highest relevance scores, thereby facilitating more accurate identification of manipulated regions within an image. The proposed methodology represents a significant advancement in deepfake detection and image forensics, by incorporating a highly effective graph-based approach with Transformer architectures that improves accuracy and enhances practical utility.

%% file: evaluation.tex
The present section endeavors to appraise the performance and efficacy of the proposed methodology through a series of challenging experiments, and subsequently compare the results against those achieved by current state-of-the-art techniques.

\subsection{Datasets} \label{sec:dataset}
\BfPara{Training datasets} 
In accordance with recent developments in the field of deepfake detection \cite{li2021frequency, li2020face, liu2021spatial, luo2021generalizing, khormali2022dfdt}, this study endeavors to evaluate the efficacy and performance of the proposed self-supervised graph Transformer model against different real-world settings through a comprehensive set of experiments. To this end, the model is trained and evaluated using several well-established and challenging deepfake forensics datasets, including FaceForensics++ \cite{rossler2019faceforensics++}, {Celeb-DF (V2)} \cite{Celeb_DF_cvpr20}, and {WildDeepfake} \cite{zi2020wilddeepfake}. A concise overview of these datasets is provided below.

\BfPara{FaceForensics++} The FaceForensics++ dataset encompasses a diverse range of four distinct types of image forgeries, specifically Deepfakes~\cite{Faceswap}, FaceSwap~\cite{Faceswap1}, Face2Face~\cite{thies2016face2face}, and NeuralTextures~\cite{thies2019deferred}. The FaceForensics++ dataset is thoughtfully curated and available in three varying compression rates, namely heavily compressed (LQ), slightly compressed (HQ), and uncompressed (Raw). While the Raw dataset exhibits relatively facile identification of deepfakes, the process becomes considerably more challenging with higher compression rates. In this manuscript, the HQ version of the dataset is predominantly utilized, unless specified otherwise.
\BfPara{Celeb-DF (V2)} The Celeb-DF (V2) dataset represents a significant advancement in synthetic video generation due to its superior face-swapping strategy that elevates its visual quality above previous efforts. The dataset encompasses an extensive collection of 5639 videos, all of which exhibit a high level of visual quality, making it an ideal resource for training and testing deepfake detection models. 
\BfPara{WildDeepfake} The WildDeepfake is a collection of fabricated videos sourced entirely from the Internet, making it a distinct and demanding real-world dataset. Unlike other AI-generated deepfakes, such as Celeb-DF (V2) and FaceForensics++, the WildDeepfake dataset incorporates multiple facial expressions and encompasses a wider range of scenarios, with numerous individuals appearing in each scene.

\BfPara{Testing datasets} In addition to the aforementioned datasets, the official test set of several additional benchmarks are utilized to investigate the generalizability of the proposed model. Specifically, these benchmarks are the DeeperForensics~\cite{jiang2020deeperforensics}, FaceShifter~\cite{li2019faceshifter}, and DeepFake Detection Challenge (DFDC) ~\cite{dolhansky2019deepfake} datasets. The DeeperForensics dataset contains over 11,000 synthetic videos generated by leveraging a specific Variational Auto-Encoder (VAE) on the real videos from FaceForensics++. Similarly, the FaceShifter dataset is a collection of videos created by applying advanced face-swapping techniques to the real videos from the same dataset. This presents a valuable opportunity to assess the robustness and effectiveness of the proposed self-supervised graph Transformer model in detecting deepfake videos across a wide range of scenarios, including those that incorporate different sophisticated manipulation techniques based on the same underlying source videos. In addition, the DFDC dataset, comprising over 4,000 manipulated videos created using various GAN-based and non-learned techniques from 1,000 real videos, provides a complementary benchmark against which the proposed method can be evaluated.

\subsection{Implementation Specifics}
\textbf{Deepfake Encoder Training.}
The present study adopts an architecture based on vision Transformers \cite{dosovitskiy2020image} as the fundamental building block of the feature extractor $f$, which is trained through the utilization of the previously mentioned contrastive learning approach expounded upon in~\textsection\ref{sec:SSL}. Specifically, the feature extractor model undergoes pre-training and fine-tuning on images of size $320 \times 320$, with patch sizes of $20 \times 20$ yielding a total of 256 patch tokens. The projection head $h$, which plays a central role in feature extraction, is established by means of a multi-layer perceptron comprising three layers. Importantly, to foster effective information sharing between the $\mathrm{CLS}$ and the patch tokens, the projection head parameters are shared, notably, ${{h}_{s}}^{[\mathrm{CLS}]} = {{h}_{s}}^{\mathrm{patch}}$ and ${{h}_{t}}^{[\mathrm{CLS}]} = {{h}_{t}}^{\mathrm{patch}}$. This enables the feature extractor model to capitalize on both the global and local context provided by the $\mathrm{CLS}$ and patch tokens, respectively. 
The training regimen is executed for 800 epochs with a learning rate that exhibits a linearly scaled progression in accordance with the batch size, as determined by the expression: $\operatorname{lr}=5 e^{-4} \times {batch_{size}} / 256$.

The preeminent implementation specifications at hand entail the pre-training of a self-supervised feature extractor on a conflation of training sets procured from three distinct benchmarks explicated in~\textsection\ref{sec:dataset}. 

Upon the conclusion of training, the parameters of the feature extractor are fixed and employed as a pre-trained feature extractor for the purpose of node feature vector extraction, which in turn facilitates the representation of the deepfake images as a graph.
The graph Transformer classifier is comprised of a graph convolutional layer, and three blocks of Transformer layers, each of which comprises 8 self-attention heads, a dimensional parameter of D=256, and an MLP size of 512 (Equation \ref{eq:attention}). This construction is indicative of the sophistication that underlies the proposed approach.

The experimental evaluation of the proposed framework was carried out on a Lambda Quad deep learning workstation machine equipped with 4 NVIDIA GeForce GTX 1080 Ti Graphics Processing Units (GPUs), 64 GB DDR4 RAM, an Intel Core™ i7-6850K CPU, and running the Ubuntu 20.04.3 LTS operating system. This machine configuration facilitated the efficient training and evaluation of the models, thanks to its high-performance computing capabilities. The performance evaluation of the proposed framework was conducted using accuracy and area under the receiver operating characteristic curve (AUC) metrics in different experimental settings. Comprehensive assessment of the proposed framework's performance was enabled through the use of these stringent evaluation metrics, which were employed to compare it with state-of-the-art methods and to provide valuable insights into the practical applicability of the approach.

%% file: results.tex
\begin{table}[!t]
\centering
\caption{Performance of the proposed self-supervised graph Transformer deepfake detection model using the in-dataset setting.}
\label{tab:IntraEvaluation}
\begin{tabular}{llcc}
\toprule
\multicolumn{2}{l}{\multirow{1}{*}{\textbf{Dataset}}}            & \textbf{ACC     (\%)}    &\textbf{ AUC    (\%)} \\ \midrule
\multirow{3}{*}{FaceForensics++}  & (Raw)  &      99.38      &  99.96                 \\
                                  & (HQ)   &      98.41      &  99.34               \\
                                  & (LQ)   &      94.59      &  95.16                \\

\multicolumn{2}{l}{Celeb-DF (V2)}         &      99.47      &   99.43            \\
\multicolumn{2}{l}{WildDeepfake}          &      81.37      &  81.24         \\ 

\bottomrule
\end{tabular}
\end{table}

To fully ascertain the proficiency of the presented deepfake detection framework, an exhaustive set of experiments was conducted, with a multifaceted approach to examine its performance across various perspectives. The scope of investigation encompassed the generalization ability of the framework in detecting deepfakes originating from heterogeneous datasets and manipulation techniques, as well as its resistance against compression and perturbations, and other ablation studies that are critical in appraising the framework's efficacy. To this end, three distinct models were meticulously trained and evaluated in an in-dataset setting using each of the preeminent deepfake forensics datasets, namely FaceForensics++, Celeb-DF (V2), and WildDeepfake, with the corresponding accuracy rate and AUC scores reported in \autoref{tab:IntraEvaluation}.

\begin{table}[t]
\centering
\caption{ Cross-dataset generalization AUC (\%) scores. The model is trained over FaceForensics++ and evaluated over a test set of DeeperForensics (DFo), FaceShifter (FSh), DeepFake Detection Challenge (DFDC), and Celeb-DF (CDF), respectively. }
\label{tab:CrossdatasetEvaluation_video}
\begin{tabular}{l|cccc|c}
\toprule
\textbf{Methods}             	&	 \textbf{DFo} 	&	 \textbf{FSh} 	&	 \textbf{DFDC} 	&	\textbf{ CDF} 	&	 \textbf{Avg.}  \\ \midrule		
Xception~\cite{rossler2019faceforensics++}      	&	84.5	&	72.0	&	70.9	&	73.7	&	 75.3 \\		
CNN-aug~\cite{wang2020cnn}                          &	74.4	&	65.7	&	72.1	&	75.6	&	 72.0 \\		
Patch-based~\cite{chai2020makes}  	                &	81.8	&	57.8	&	65.6	&	69.6	&	 68.7 \\		
Face X-ray~\cite{li2020face}    	                &	86.8	&	92.8	&	65.5	&	79.5	&	 81.2 \\		
CNN-GRU~\cite{sabir2019recurrent}      	            &	74.1	&	80.8	&	68.9	&	69.8	&	 73.4 \\		
Multi-task~\cite{nguyen2019multi}  	                &	77.7	&	66.0	&	68.1	&	75.7	&	 71.9 \\		
DSP-FWA~\cite{li2019exposing}       	            &	50.2	&	65.5	&	67.3	&	69.5	&	 63.1 \\		
LipForensics~\cite{haliassos2021lips} 	            &	97.6	&	97.1	&	73.5	&	82.4	&	 87.7 \\		
FTCN \cite{zheng2021exploring}       	            &	98.8	&	98.8	&	74.0	&	86.9	&	 89.6 \\		
DFDT  \cite{khormali2022dfdt}                	    &	96.9	&	97.8	&	76.1	&	88.3	&	 89.7 \\ 		
RealForensics \cite{haliassos2022leveraging}       	&	99.3	&	99.7	&	75.9	&	86.9	&	 90.5 \\\midrule		
Ours                   	                            &	98.9	&	99.1	&	77.3	&	87.9	&	 90.8 \\ 		\bottomrule
\end{tabular}
\end{table}

\subsection{Cross Dataset Generalization}
The proliferation of fake videos in the real world poses a formidable challenge for the deployment of deepfake detection models, given the diverse forgery techniques, heterogeneous source videos, and extensive post-processing techniques. Therefore, it is crucial to assess the effectiveness of any deployed deepfake detection model from a domain generalization perspective. Although evaluating domain generalization is complicated due to significant disparities between deepfake forensics datasets, this can be addressed through cross-dataset evaluation. This study adopts a similar approach to the existing literature by utilizing the training set of the FaceForensics++ dataset to construct a deepfake detection model and then evaluate its performance on other datasets. More specifically, the model is trained on four distinct types of forgeries that make up the FaceForensics++ dataset. The model's performance is then evaluated on the test sets of four separate benchmark deepfake detection datasets, including DeeperForensics \cite{jiang2020deeperforensics}, FaceShifter \cite{li2019faceshifter}, Deepfake Detection Challenge \cite{dolhansky2019deepfake}, and Celeb-DF (V2) \cite{Celeb_DF_cvpr20}. This evaluation process aims to determine the model's ability to generalize across different forgery techniques and datasets, thus assessing its domain generalization capabilities. It is noteworthy that this evaluation setup is exceedingly arduous since the original and forged videos in DFDC and Celeb-DF (V2) are different from those in FaceForensics++ and were not seen during the training process. 

The results obtained from this cross-dataset evaluation are presented in \autoref{tab:CrossdatasetEvaluation_video}. Based on the results shown in the table, it is evident that the proposed graph Transformer model outperforms all other methods in terms of average AUC scores, achieving an impressive score of 90.8\%. The RealForensics method comes in second place with an average AUC score of 90.5\%, while other methods have lower scores, ranging from 63.1\% to 89.7\%. It is worth noting that some methods, such as Face X-ray and DSP-FWA, perform well on specific datasets but have lower scores on others. This indicates that their performance may be dataset-dependent and may not generalize well across different datasets. On the other hand, the presented method in this study and some other methods, such as FTCN, DFDT, and RealForensics, achieve consistently high scores across all datasets, indicating that they are robust and generalize well across different datasets. In particular, the presented method achieves the highest AUC score on the DeeperForensics, FaceShifter, and Celeb-DF (V2) datasets and the second-highest score on the DFDC dataset. It is pertinent to mention that due to the fact that DeeperForensics and FaceShifter utilize the same source videos as FaceForensics++, they tend to perform better than DFDC and Celeb-DF (V2) in terms of deepfake detection. This is because the models have already been trained on similar data, making it easier for them to generalize to these datasets.

\begin{table}[!t]
\centering
\caption{Cross-manipulation generalization results on FaceForensics++ dataset. The AUC scores (in percentage) are for each forgery class on the FaceForensics++ dataset after training on the other three manipulation types.}
\label{tab:crossmanipulation}
\begin{tabular}{l|cccc|c}
\toprule
\multirow{2}{*}{\textbf{Methods} }        & \multicolumn{4}{c|}{\textbf{Train on remaining sets}} & \multicolumn{1}{l}{\multirow{2}{*}{\textbf{Avg.}}} \\ \cline{2-5}
                               & \textbf{DF}           & \textbf{NT}           & \textbf{F2F } & \textbf{FS}       & \multicolumn{1}{l}{}                      \\ \midrule
Xception~\cite{rossler2019faceforensics++} 	               	&	93.9	&	79.7	&	86.8	&	51.2	&	 77.9                                      \\		
CNN-aug~\cite{wang2020cnn}                       	     	&	87.5	&	67.8	&	80.1	&	56.3	&	 72.9                                      \\		
Patch-based~\cite{chai2020makes}                  		    &	94.0	&	84.8	&	87.3	&	60.5	&	 81.7                                      \\		
Face X-ray~\cite{li2020face}                   		        &	99.5	&	92.5	&	94.5	&	93.2	&	 94.9                                      \\		
CNN-GRU~\cite{sabir2019recurrent}                      		&	97.6	&	86.6	&	85.8	&	47.6	&	 79.4                                      \\		
LipForensics~\cite{haliassos2021lips}                 		&	99.7	&	99.1	&	 {99.7}         	&	90.1	&	 97.1                                      \\ 		
DFDT\cite{khormali2022dfdt}                           		&	99.8	&	99.2	&	99.6	&	93.1	&	   97.9                                 \\		
AV DFD \cite{zhou2021joint}                           		&	100.0 	&	98.3	&	99.8	&	90.5	&	   97.1                                 \\		
FTCN \cite{zheng2021exploring}                        		&	99.9	&	99.2	&	99.7	&	99.9	&	  99.6                                 \\ 		
RealForensics \cite{haliassos2022leveraging}                &	100.0 	&	99.2	&	99.7	&	97.1	&	  99.6                                 \\ \midrule		
Ours                                                   		&	99.9	&	99.2	&	99.8	&	98.3	&	   99.3                                 \\		\bottomrule
\end{tabular}
\end{table}

\subsection{Cross Manipulation Generalization}
In the real world, deepfakes can be generated using various techniques, including different source videos and post-processing methods, resulting in a vast array of manipulated videos. A deepfake detection model that is only trained on a specific dataset may fail to detect deepfakes generated using different techniques, rendering the model ineffective. Therefore, any deployed deepfake detection algorithm must perform well on unseen forgery types \cite{haliassos2021lips, li2020face, nguyen2019multi, zheng2021exploring}. This means that cross-manipulation generalization is critical for deepfake detection as it enables the detection model to perform effectively on unseen and potentially more sophisticated manipulated videos.   
To this end, the cross-manipulation generalization property of the presented approach is investigated using the leave-one-out strategy \cite{haliassos2021lips, khormali2022dfdt, masi2020two, chai2020makes, nguyen2019multi}. Specifically, model performance is evaluated on each forgery type in FaceForensics++ dataset after training with the remaining manipulation techniques. The obtained results on the test sets of Deepfakes (DF),  NeuralTextures (NT), Face2Face (F2F), and FaceSwap (FS) are shown in \autoref{tab:crossmanipulation}. The table shows that the proposed method achieves a high average AUC score of 99.3\%, which is comparable to state-of-the-art methods, such as FTCN and RealForensics. The high average AUC score indicates that the presented method's cross-manipulation extends well to previously unseen manipulation types. It works on par with the SOTA with a self-supervised feature extractor, indicating its effectiveness in capturing more generalizable features for accurate deepfake detection.

\begin{table}[t]
\centering
\caption{Generalization AUC (\%) scores across different compression levels. The model is trained over Neuraltextures (NT) forgery type and tested on Deepfakes (DF) and FaceSwap (FS) forgery types from the FaceForensics++ dataset at two compression levels, \ie LQ and HQ. }
\label{tab:crosscompression}
\begin{tabular}{l|cc|cc}
\toprule
\multirow{2}{*}{\textbf{Method}}  & \multicolumn{2}{c|}{\textbf{HQ}} & \multicolumn{2}{c}{\textbf{LQ}} \\ \cline { 2 - 5}
                        & \textbf{FS}         & \textbf{DF}        & \textbf{FS}         & \textbf{DF}        \\ \midrule
Xception~\cite{rossler2019faceforensics++}               & 71.8      & 77.0     & 51.7      & 58.7     \\
Face X-ray~\cite{li2020face}             & 77.9      & 58.5     & 51.0      & 57.1     \\
F3Net \cite{qian2020thinking}                  & 61.2      & 80.5     & 51.9      & 58.3     \\
RFM  \cite{wang2021representative}                   & 63.9      & 79.8     & 51.6      & 55.8     \\
SRM  \cite{luo2021generalizing}                   & 79.5      & 83.8     & 52.9      & 55.5     \\
SLADD \cite{chen2022self}               & 72.1      & 84.6     & 56.8      & 62.8     \\ \midrule
Ours                    &  72.9     & 84.5     & 57.2      &  61.4    \\ 

\bottomrule
\end{tabular}
\end{table}

\begin{figure}[t]
    \centering
    \includegraphics[width=0.5\textwidth]{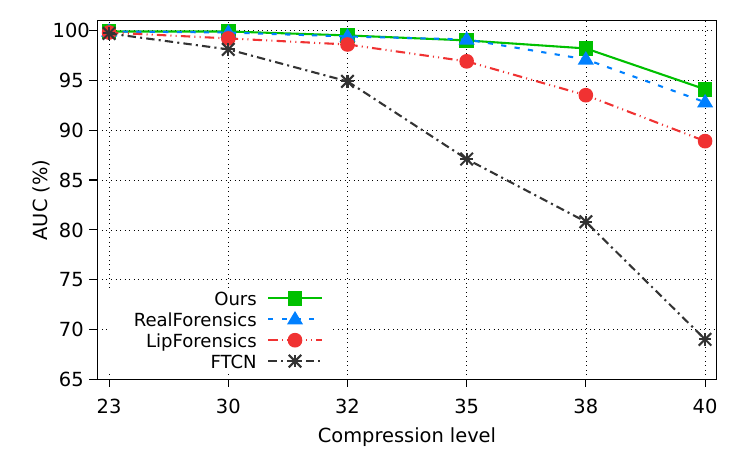}
    \caption{Robustness of the model against different compression levels. Models are trained over the FaceForensics++ dataset with a compression level of 23 and evaluated against five different compression levels. }
    \label{fig:compression}
\end{figure}

\begin{table*}[t]
\centering
\caption{Robustness against common post-processing perturbations. The experiments are performed using five different intensity levels for each perturbation type and reported the average AUC score (\%). Moreover, the average AUC score across all corruptions for each deepfake detection method is presented. }
\label{tab:perturbation}
\begin{tabular}{l|ccccccc|c}
\toprule
\textbf{Method}                & \textbf{Clean} & \textbf{Saturation} & \textbf{Blur} & \textbf{Block} & \textbf{Contrast} & \textbf{Pixel} & \textbf{Noise} & \textbf{Avg}  \\ \midrule
Xception~\cite{rossler2019faceforensics++}    & 99.8  & 99.3       & 60.2 & 99.7  & 98.6     & 74.2  & 53.8  & 78.3 \\
CNN-aug~\cite{wang2020cnn}     & 99.8  & 99.3       & 76.5 & 95.2  & 99.1     & 91.2  & 54.7  & 84.1 \\
Patch-based~\cite{chai2020makes}  & 99.9  & 84.3       & 54.4 & 99.2  & 74.2     & 56.7  & 50.0  & 67.5 \\
Face X-ray~\cite{li2020face}   & 99.8  & 97.6       & 63.8 & 99.1  & 88.5     & 88.6  & 49.8  & 77.5 \\
CNN-GRU~\cite{sabir2019recurrent}      & 99.9  & 99.0       & 71.5 & 97.9  & 98.8     & 86.5  & 47.9  & 82.3 \\
LipForensics~\cite{haliassos2021lips}  & 99.9  & 99.9       & 96.1 & 87.4  & 99.6     & 95.6  & 73.8  & 92.5 \\
FTCN \cite{zheng2021exploring}         & 99.4  & 99.4       & 95.8 & 97.1  & 96.7     & 98.2  & 53.1  & 89.5 \\ 
RealForensics \cite{haliassos2022leveraging} & 99.8  & 99.8       & 95.3 & 98.9  & 99.6     & 98.4  & 79.7  & 95.6 \\\midrule
Ours  & 99.7  & 99.8       & 96.1 & 98.8  & 99.7     & 98.2  & 81.1  & 96.2 \\

\bottomrule
\end{tabular}
\end{table*}

\subsection{Cross Compression Generalization}
In addition to achieving high generalization scores across different types of deepfake manipulations, ensuring the robustness of the deployed detector against varying levels of compression is equally crucial. The proposed method's cross-compression generalization is evaluated through testing on Deepfakes and FaceSwap forgery types with different compression levels, \ie LQ and HQ, while being trained on NeuralTextures from the FaceForensics++ dataset. The results obtained are presented in Table \ref{tab:crosscompression}. It can be observed that while competitive generalization was demonstrated by the proposed graph Transformer model when the compression level was kept constant during training and testing. However, a significant drop in performance was noted once tested against compressed samples, as evidenced in Figure \ref{fig:compression}. This is unsurprising, given that highly compressed images are subject to the loss of textural features and low-level clues. Therefore, FTCN and LipForensics were highly affected by compression levels. However, the proposed self-supervised graph transformer model achieves slightly better performances as it leverages high-level facial representations, which are less susceptible to low-level artifacts.

\subsection{Resilience to Perturbation}
Given that real-world forgeries are vulnerable to various forms of corruption on social media, such as saturation, contrast, blur, etc., it is essential that the deployed deepfake detector can withstand such post-processing perturbations. The procedure proposed in \cite{haliassos2021lips} is taken into account to investigate the resilience of the model against common post-processing filters, including applying Gaussian noise \& blur, modifying saturation \& contrast, pixelation, and block-wise occlusions \cite{jiang2020deeperforensics}. The model is trained over grayscale clips of the FaceForensics++ dataset, with only horizontal flipping and random cropping augmentations, and evaluated against five different intensity levels of aforementioned perturbations. The resulting average AUC scores are listed in \autoref{tab:perturbation}. The presented graph Transformer model is significantly less vulnerable to aforementioned corruptions compared to other methods that utilize low-level cues such as Patch-based and Face X-ray. Furthermore, the proposed method outperforms FTCN and LipForensics, while exhibiting comparable results to RealForensics, thus showcasing its competitive performance.

\begin{table*}[!h]
\centering
\caption{A quantitative comparison of methods performance on every dataset with existing deepfake detection approaches in frame-level analysis. Reported results are obtained from associated articles. The~same evaluation metric as the literature is used for each dataset to provide a fair comparison and better insight into the model's~performance.} \label{tab:IntraEvaluation_Comparison}
\begin{tabular}{lc|lc|lc}
\toprule
\multirow{2}{*}{\textbf{Methods} } & \textbf{FaceForensics++} & \multirow{2}{*}{\textbf{Methods} } & \multirow{1}{*}{\textbf{Celeb-DF (V2)}} & \multirow{2}{*}{\textbf{Methods} } & \multirow{1}{*}{\textbf{WideDeepfake}}                                      \\ 
                & \textbf{AUC (\%)}        &                         &\textbf{ AUC (\%)}                       &                         & \textbf{(ACC \%)}                                                            \\ \midrule
	Two-stream~\cite{zhou2017two}	                &	70.1	&	Two-stream~\cite{zhou2017two}	                    &	53.8	&	AlexNet~\cite{krizhevsky2017imagenet}	        &	60.3	\\	
	Meso4~\cite{afchar2018mesonet}	                &	84.7	&	Meso4~\cite{afchar2018mesonet}	                    &	54.8	&	VGG16~\cite{simonyan2015very}	                &	60.9	\\	
	HeadPose~\cite{yang2019exposing}	            &	47.3	&	HeadPose~\cite{yang2019exposing}	                &	54.6	&	ResNetV2-50~\cite{he2016identity}	            &	63.9	\\	
	FWA~\cite{li2019exposing}	                    &	80.1	&	FWA~\cite{li2019exposing}	                        &	56.9	&	ResNetV2-101~\cite{he2016identity}	            &	58.7	\\	
	VA-MLP~\cite{matern2019exploiting}	            &	66.4	&	VA-MLP~\cite{matern2019exploiting}	                &	55.0	&	ResNetV2-152~\cite{he2016identity}	            &	59.3	\\	
	Xception~\cite{rossler2019faceforensics++}	    &	99.7	&	Xception~\cite{rossler2019faceforensics++}	        &	65.5	&	Inception-v2~\cite{szegedy2016rethinking}	    &	62.1	\\	
	Multi-task~\cite{nguyen2019multi}	            &	76.3	&	Multi-task~\cite{nguyen2019multi}	                &	54.3	&	MesoNet-1~\cite{afchar2018mesonet}	            &	60.5	\\	
	Capsule~\cite{nguyen2019use}	                &	96.6	&	Capsule~\cite{nguyen2019use}	                    &	57.5	&	MesoNet-4~\cite{afchar2018mesonet}	            &	64.4	\\	
	DSP-FWA~\cite{li2019exposing}	                &	93.0	    &	DSP-FWA~\cite{li2019exposing}	                    &	64.6	&	MesoNet-inception~\cite{afchar2018mesonet}	    &	66.0	\\	
	Two-branch~\cite{masi2020two}	                &	93.2	&	Two-branch~\cite{masi2020two}	                    &	73.4	&	XceptionNet~\cite{chollet2017xception}	        &	69.2	\\	
	SPSL~\cite{liu2021spatial}	                    &	96.9	&	Face X-ray~\cite{li2020face}	                    &	80.5	&	ADDNet-2D~\cite{zi2020wilddeepfake}	            &	76.2	\\	
 	F3-Net~\cite{qian2020thinking}	                &	97.9	&	SPSL~\cite{liu2021spatial}	                        &	76.8	&	ADDNet-3D~\cite{zi2020wilddeepfake}	            &	65.5	\\	
	Video SE~\cite{khan2021video}		            &	99.6   &	F3-Net~\cite{qian2020thinking}	                    &	65.1	&	ADD-Xception~\cite{khormali2021add}	            &	79.2	\\	
	RNN~\cite{guera2018deepfake}                    &	83.1   &	PPA~\cite{charitidis2020investigating}	            &	83.1	&	DFDT~\cite{khormali2022dfdt}                    &	81.3    	\\	
	DFDT~\cite{khormali2022dfdt}                    &	99.7   &	DefakeHop~\cite{chen2021defakehop}	                &	90.5	&			                                        &	    	\\	
		                                            &	    	&	FakeCatcher~\cite{ciftci2020fakecatcher}	        &	91.5	&	                                    	        &	    	\\	
	                                                &		    &	ATS-DE~\cite{tran2021high}	                        &	97.8	&			                                        &	    	\\	
	                                                &		    &	ADD-ResNet~\cite{khormali2021add}	                &	98.3	&			                                        &	    	\\
	                                                &           &   DFDT~\cite{khormali2022dfdt}                        &99.2 &  &   \\  	\midrule
	
	Ours                                           &	{99.9}	    &	Ours              	                &	  {99.4}  	&   Ours			                    &	    {81.3	}\\	
\bottomrule
\end{tabular}
\end{table*}

\subsection{Comparison with SOTA}
A comprehensive evaluation was carried out to assess the superiority of the presented deepfake detection framework over state-of-the-art models. In particular, the model's performance was compared against state-of-the-art deepfake detection models that were trained and tested using same datasets, and the resultant outcomes are presented in \autoref{tab:IntraEvaluation_Comparison}. Remarkably, the model exhibited competitive performance when evaluated on the FaceForensics++ dataset, while outperforming other models when tested on Celeb-DF (V2) and WildDeepfake datasets. The obtained results unequivocally highlight the exceptional potency of the proposed graph Transformer model, underscored by its superlative generalizability attributable to the high-level deepfake representations obtained by a self-supervised framework. Furthermore, the proposed model's ability to leverage the underlying graph structure of the deepfake images enables the identification of both local and global subtle discrepancies and artifacts that may evade detection using traditional methods. This highlights the significance of the proposed model, which offers a compelling alternative to existing deepfake detection methods and has the potential to significantly enhance the reliability and robustness of deepfake detection systems.

\subsection{Ablation Studies and Model Configuration}

The proposed framework was comprehensively evaluated through a series of meticulous ablation studies, aiming to investigate the role of different components of the model and its hyperparameter values. The outcomes of these experiments on Celeb-DF (V2) are summarized in Table \ref{tab:KImpact}. Firstly, the impact of various feature extractors on the overall performance of the classification task was assessed. The combination of visual representations obtained from the proposed constructive learning vision Transformer architecture with the graph Transformer classifier exhibited superior performance compared to utilizing a pre-trained feature extractor, specifically ResNet50, in combination with the same graph Transformer classifier. 

A series of experiments were conducted using both sets of feature extractors to investigate different hyperparameters' impacts. For example, the influence of the number of included neighboring patches, denoted as $K$, on the graph construction process and the overall efficacy of the deepfake detection task is investigated. Table \ref{tab:KImpact} reveals that the best performance was achieved when $K=8$. Smaller values of $K$ yield a reduction in the receptive field of the constructed graph, consequently resulting in lower performance. Lastly, the impact of other critical hyperparameters, such as the number of graph convolutional layers and the number of Transformer blocks, on the overall performance of the deepfake detection pipeline is investigated. The obtained results demonstrated a slight degradation in performance when the number of graph convolution layers decreased from three layers to one layer. This observation can be attributed to the concept of node receptive field, where a greater number of graph convolution layers allows for the incorporation of long-term information during model development. Moreover, it was observed that utilizing three Transformer blocks facilitated more effective pooling of discriminative information, leading to improved performance in deepfake detection tasks.

The conducted ablation studies collectively showcased the superior performance of the proposed self-supervised graph Transformer framework for deepfake detection. These findings contribute to the advancement of the field and emphasize the efficacy of the employed methodology for addressing the challenges associated with deepfake detection.

\begin{table}[]
\centering
\caption{Performance evaluation of the proposed self-supervised graph Transformer deepfake detection model using different ablation studies on Celeb-DF (V2) dataset. Key variables include the feature extractor (FE), neighboring patches count ($K$), graph convolutional layers (GCN), and number of Transformer blocks (TB).}
\label{tab:KImpact}
\begin{tabular}{c|c|c|c|c}
\toprule
\textbf{FE}                     & \textbf{K}  & \textbf{GCN}                & \textbf{TB} & \textbf{AUC (\%)} \\ \midrule
\multirow{8}{*}{ResNet50} & \multirow{4}{*}{4} & \multirow{2}{*}{1} & 3                              &   97.7         \\
                                      &                    &                    & 6                  &   98.1      \\
                                      &                    & \multirow{2}{*}{3} & 3                  &   97.7       \\
                                      &                    &                    & 6                  &   97.8       \\ \cline{2-5}
                                      & \multirow{4}{*}{8} & \multirow{2}{*}{1} & 3                  &   98.4       \\
                                      &                    &                    & 6                  &   98.3       \\
                                      &                    & \multirow{2}{*}{3} & 3                  &   98.5       \\
                                      &                    &                    & 6                  &   98.4       \\ \midrule
\multirow{8}{*}{SSL ViT}              & \multirow{4}{*}{4} & \multirow{2}{*}{1} & 3                  &   98.6       \\
                                      &                    &                    & 6                  &   98.6       \\
                                      &                    & \multirow{2}{*}{3} & 3                  &   99.1       \\
                                      &                    &                    & 6                  &   98.9        \\ \cline{2-5}
                                      & \multirow{4}{*}{8} & \multirow{2}{*}{1} & 3                  &   98.7        \\
                                      &                    &                    & 6                  &   98.6       \\
                                      &                    & \multirow{2}{*}{3} & 3                  &   99.4       \\
                                      &                    &                    & 6                  &   99.1      \\ \bottomrule
\end{tabular}
\end{table}

%% file: conclusion.tex
This work presents a self-supervised graph Transformer framework for deepfake detection that is comprised of three main components. The proposed framework leverages the expressive power of self-supervised contrastive learning to learn high-level representations of deepfakes. Unlike existing supervised approaches, which depend on low-level visual fingerprints, the extracted high-level visual representations via this approach significantly enhance the generalizability of deepfake detection models, thereby reducing their dependence on dataset-specific patterns. Furthermore, the current work leverages the expressive power of graph convolutional networks and Transformer blocks to capture intricate interdependencies among both local and global regions of an image, rendering them particularly suited to deepfake detection tasks. In addition, the proposed graph Transformer relevancy map provides a comprehensive understanding of the complex interdependencies and salient features which leads to enhanced transparency and accountability of the detection model. Finally, the efficacy and generalizability of the proposed framework were rigorously evaluated via a comprehensive set of experiments, encompassing a diverse range of challenging scenarios spanning both in-distribution and out-of-distribution settings. The experimental results unequivocally establish the framework's superiority, with exceptional in-dataset detection accuracy being achieved. Moreover, the proposed self-supervised pre-training feature extractor constitutes a significant contribution to the field, having markedly improved the framework's ability to generalize across multiple datasets while simultaneously enhancing its resilience to post-processing perturbations, such as compression and blur.